\documentclass[letterpaper, 10 pt, conference]{ieeeconf}  
\usepackage{graphicx}
\usepackage{booktabs}
\usepackage{subcaption}
\usepackage{mathrsfs}
\usepackage{mathtools}
\usepackage{amssymb}

\IEEEoverridecommandlockouts                              

\usepackage{graphics} 
\usepackage{amsmath} 

\title{\LARGE \bf
Lights Out: A Nighttime UAV Localization Framework Using Thermal Imagery and Semantic 3D Maps
}

\author{Ryan Allen and Melissa Greeff
\thanks{The authors are with Robora Lab (www.roboralab.com),
Queen’s University; and affiliated with Ingenuity Labs Research
Institute. E-mails: 19rca4@queensu.ca,
melissa.greeff@queensu.ca.}}

\begin{document}

\bstctlcite{BSTcontrol}
\maketitle
\thispagestyle{empty}
\pagestyle{empty}

\begin{abstract}
Reliable backup localization for unmanned aerial vehicles (UAVs) operating in GNSS-denied nighttime conditions remains an open challenge due to the severe modality gap between daytime RGB maps and nighttime thermal imagery. This work presents a semantic reprojection framework for map-relative nighttime UAV localization by aligning segmented thermal observations with a globally referenced, semantically labeled 3D map constructed from daytime RGB data. Rather than relying on appearance-based correspondence, localization is formulated in a shared semantic domain and solved via a symmetric bidirectional reprojection objective with confusion-aware weighting to improve robustness under segmentation uncertainty. The approach is evaluated offline across  6.5\,km of nighttime, real-world UAV flight trajectories in urban and semi-structured environments. Relative to RTK GNSS ground truth, the system achieves a bias-corrected $\mathrm{RMSE}_{2D}$ of 2.18\,m and a median $\mathrm{RMSE}_{2D}$ of 1.52\,m. Results show that localization performance is strongly correlated with the availability of semantic edge evidence and that large-error events are spatially localized to semantically ambiguous areas rather than uniformly distributed. These findings indicate that semantic reprojection offers a promising pathway toward globally referenced nighttime UAV localization using thermal imagery alone.
\end{abstract}

\begin{figure}
  \centering
  \includegraphics[width=\linewidth]{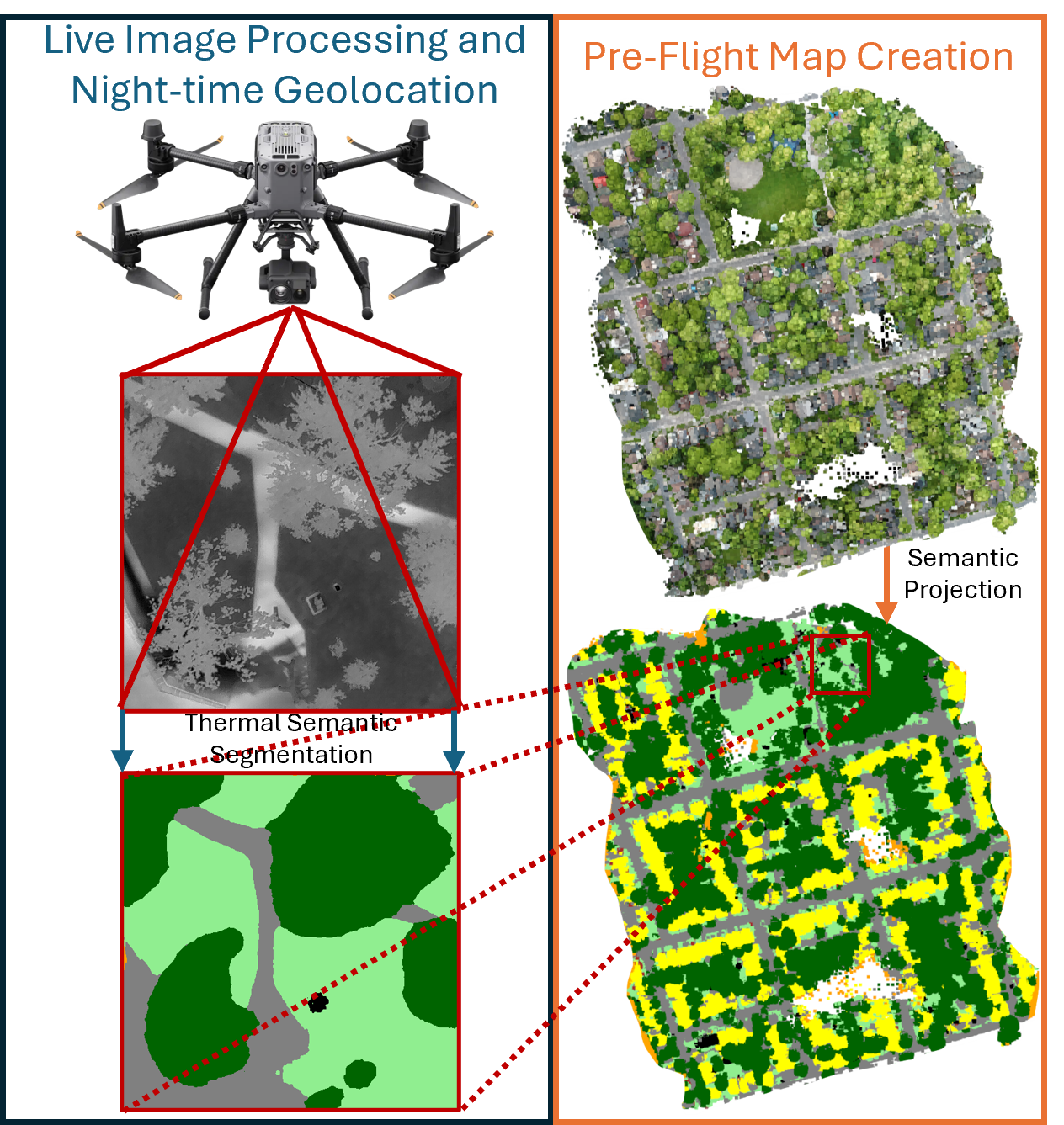}
  \caption{\small Overview of the proposed nighttime localization pipeline with the Vic dataset. A semantically labeled 3D map is constructed from daytime RGB imagery during pre-flight mapping (right). At night, segmented thermal imagery (left) is aligned to this map via edge-aware reprojection, enabling globally referenced localization without appearance-based matching.}
  \label{fig:small_diagram}
\end{figure}
\section{Introduction}

\begin{figure*}[t]
    \centering
    \includegraphics[width=\textwidth]{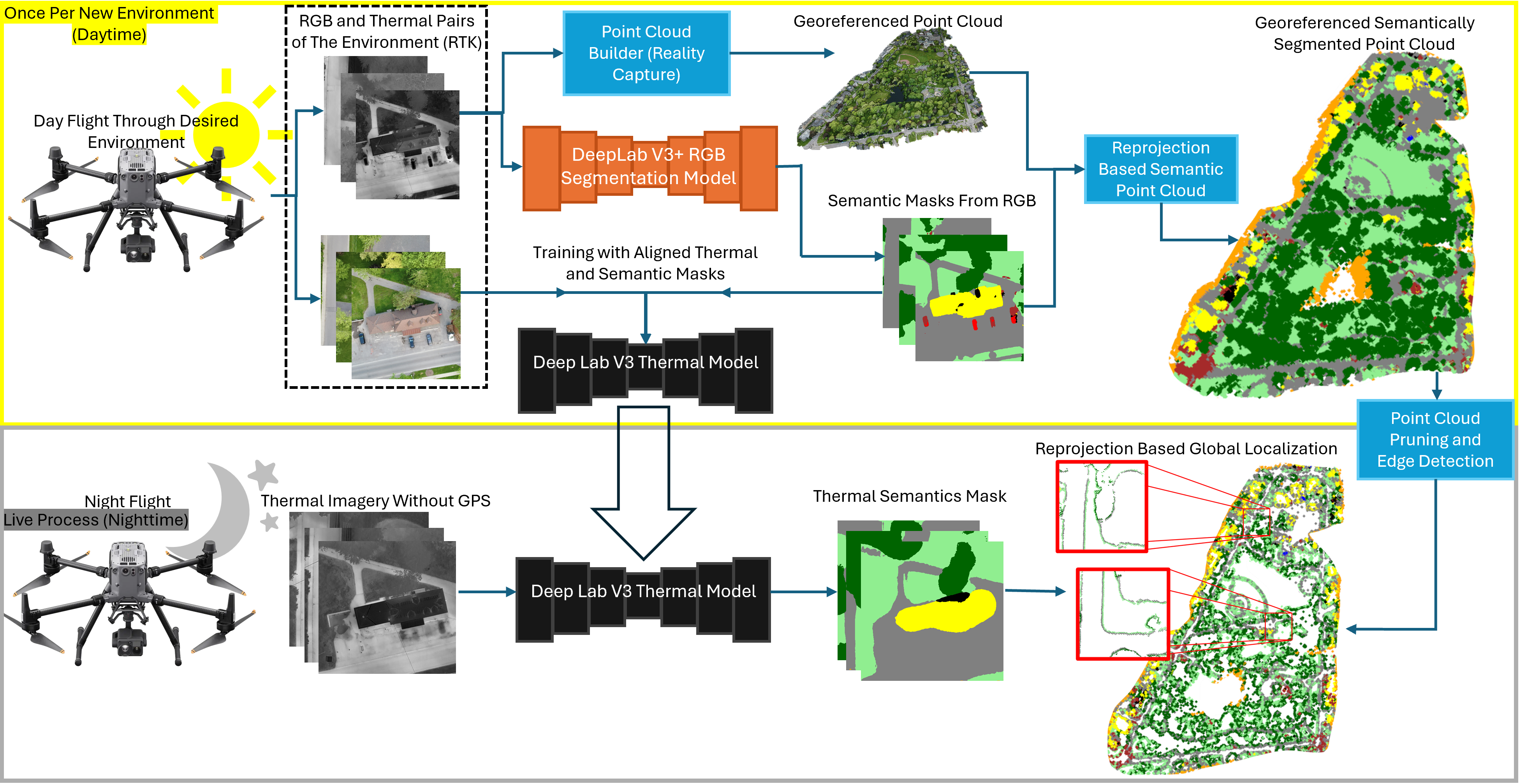}
    \caption{\small Overview of the proposed thermal semantic localization pipeline with the City dataset. \textbf{Daytime preprocessing} (top): synchronized RGB–thermal data with RTK ground truth are used to train a thermal segmentation model and construct a georeferenced point cloud. Semantic labels are reprojected into the 3D reconstruction to produce a semantically labeled map, which is refined through pruning and edge extraction. \textbf{Nighttime operation} (bottom): incoming thermal images are segmented and aligned with the semantic 3D map via reprojection, enabling globally referenced, map-relative localization under nighttime conditions.
}
    \label{fig:pipeline}
\end{figure*}
Autonomous navigation of unmanned aerial vehicles (UAVs) is expanding into safety-critical domains including parcel delivery \cite{lingam_human_2025}, infrastructure inspection \cite{chodura_cold_2025}, and public safety response, where reliable global localization is essential. In practice, most UAV navigation stacks depend on GNSS as the primary source of global reference, using visual–inertial odometry (VIO) or IMU-based dead reckoning for short-term stabilization. However, GNSS signals are vulnerable to occlusion, multipath interference, jamming, and spoofing, making exclusive reliance on satellite positioning unacceptable for many safety-critical deployments.

UAV localization methods broadly fall into two categories: (1) online SLAM, which estimates pose while incrementally building a map, and (2) map-relative localization, which aligns onboard observations to a pre-built global reference map \cite{warren_theres_2019, bianchi_uav_2021}. Under daylight conditions, both paradigms perform reliably because visual–inertial pipelines can exploit stable RGB texture, photometric gradients, and appearance-based feature correspondences \cite{xu_slam_2025}.

At night, however, these cues degrade dramatically. RGB-based registration to daytime maps becomes unreliable because feature detection and data association assume illumination-consistent appearance \cite{patel_visual_2020}. This exposes a fundamental modality mismatch: global maps are typically constructed from RGB satellite imagery or RGB structure-from-motion (SfM) reconstructions \cite{jiang_efficient_2020}, while nighttime UAV operation frequently relies on thermal sensing, which encodes scene structure through heat patterns rather than color or texture.

Alternative sensing modalities provide only partial solutions. LiDAR offers illumination-invariant geometric registration but introduces substantial size, weight, and power (SWaP) penalties that limit scalability for many UAV platforms \cite{storch_comparative_2025}. Thermal-inertial odometry and thermal SLAM methods mitigate lighting dependence but accumulate drift over long trajectories and remain challenging in large-scale outdoor environments \cite{xu_slam_2025, wu_monocular_2025, jiang_thermal-inertial_2022}. Multi-spectral RGB–thermal fusion improves perception robustness but typically assumes concurrent visible-spectrum sensing and does not address operation in complete darkness \cite{brenner_rgb-d_2023}.

This work addresses the nighttime localization problem by reformulating map-relative UAV localization as a semantic reprojection problem. Instead of relying on appearance similarity, we align thermal observations and daytime RGB maps in a shared semantic domain. A semantically labeled 3D map is constructed offline from daytime RGB data. During nighttime flight, thermal imagery is segmented and aligned to this map via semantic reprojection in the global frame. By operating on semantic structure rather than photometric texture, the method bridges the RGB--thermal modality gap and enables structural matching across sensing domains. Unlike pixel-level alignment methods, semantic alignment is invariant to illumination changes, contrast inversion, and radiometric differences between RGB and thermal modalities, enabling consistent structural correspondence without requiring photometric agreement.

Unlike prior nighttime localization approaches that either (i) rely on thermal SLAM or (ii) align thermal imagery to 2D satellite maps via appearance-based descriptors and homography estimation, we perform \emph{cross-modal, globally referenced alignment in a persistent 3D semantic map}. Nighttime thermal observations are aligned by projecting this map into the thermal image under candidate global poses, and localization is formulated as a \emph{symmetric semantic reprojection objective} with confusion-aware marginalization. This formulation replaces appearance consistency with structural semantic consistency enforced directly in the global 3D frame and enables GNSS-denied nighttime localization using thermal imagery alone at inference time. The contributions of this work are three-fold:
\begin{enumerate}
    \item We introduce a map-relative nighttime localization formulation that aligns thermal semantic edges to a globally referenced semantically labeled 3D map via a bidirectional (symmetric) semantic reprojection loss. The method enables global pose correction without appearance-based matching or LiDAR at inference time.
    \item We propose a confusion-marginalized weighting of the semantic loss objective that explicitly accounts for thermal segmentation ambiguity, significantly reducing large-error outliers under nighttime degradation.
    \item Across 6.5\,km of nighttime flight (1,373 frames), the system achieves a 2D root-mean-square error (RMSE) of 2.18\,m relative to RTK ground truth. We further demonstrate that localization error is strongly correlated with semantic edge density and that large-error events are spatially localized.
\end{enumerate}

\section{Related Work}

\subsection{Localization}

\subsubsection{Visual and LiDAR SLAM in Low-Light Environments}

Visual SLAM systems achieve high accuracy in well-illuminated environments by exploiting photometric texture and feature correspondences. Methods such as ORB-SLAM3, Ground-VIO \cite{zhou_ground-vio_2024}, and tightly coupled visual–inertial systems including VINS-Mono \cite{luo_supervins_2025} demonstrate strong performance in daylight but degrade substantially as illumination decreases due to reduced feature contrast and unstable data association.

LiDAR-based systems circumvent illumination dependence by relying on geometric registration. Approaches such as LOAM \cite{zhang_loam_2014}, LeGO-LOAM \cite{shan_lego-loam_2018}, LIO-SAM \cite{shan_lio-sam_2020}, and FAST-LIO2 \cite{xu_fast-lio2_2022} provide lighting-invariant localization but have limited scalability for lightweight UAV platforms. While effective, these systems rely on active sensing rather than passive thermal imaging.

\subsubsection{Thermal-Based Localization}

Thermal sensing provides illumination-independent observations and has been explored for nighttime navigation. Thermal-inertial odometry and SLAM methods \cite{delaune_thermal-inertial_2019,li_wti-slam_2025,lv_visual-inertial_2024} enable short-range state estimation but accumulate drift over long trajectories, limiting their ability to provide persistent global reference.

Recent thermal–satellite geo-localization approaches \cite{xiao_long-range_2023,xiao_uasthn_2025} focus on aligning UAV thermal imagery with 2D satellite maps using learned cross-domain descriptors and deep homography estimation. These methods are primarily designed for mid- to high-altitude flight regimes where UAV imagery approximates nadir satellite views and global position is inferred through image-to-image matching within a bounded search region. In contrast, our work targets lower-altitude UAV operation and aligns thermal observations to a persistent semantically labeled 3D map constructed from daytime RGB data. Rather than estimating 2D homographies based on appearance similarity, we enforce structural consistency through semantic reprojection in a global 3D frame.

\subsection{Semantic Segmentation}

Semantic segmentation provides an illumination-robust representation of scene structure and has been widely studied for both RGB and thermal imagery. Cross-modal approaches such as HeatNet \cite{vertens_heatnet_2020} demonstrate that semantic features remain more stable across sensing domains than raw appearance, enabling supervision transfer between RGB and thermal data. Subsequent work explores RGB–thermal perception using dual-camera systems and synthetic data generation \cite{meng_robust_2022,ji_uav_2025,upadhyay_comprehensive_2025}. However, these works focus primarily on segmentation accuracy and perception tasks rather than globally referenced localization.

\subsection{Semantic Representations for Localization}

Semantic representations have been incorporated into localization pipelines to improve robustness under lighting and seasonal variation, particularly in ground robotics \cite{cramariuc_semsegmap_2021,zhang_cross-modal_2023,li_vision_2020}. These approaches often combine semantics with RGB or LiDAR sensing and operate within a single sensing modality.

\begin{table*}[t]
    \centering
    \caption{Summary of Flight Operations Across Test Sites}
    \label{tab:flight_paths}
    \begin{tabular}{lcccccccc}
        \toprule
        \textbf{Location} & \textbf{Photos} & \textbf{Duration} & \textbf{Distance} & \textbf{Above Ground} & \textbf{Date} & \textbf{Day Start} & \textbf{Night Start} & \textbf{Sunset} \\
         & \textbf{Taken} & \textbf{(min)} & \textbf{(m)} & \textbf{Level (m)} & & \textbf{Time} & \textbf{Time} & \textbf{Time} \\
        \midrule
        Pier         & 542  & 21 & 2579  & 150     & 03-Jul-25 & 5:27 PM & 10:18 PM & 8:54 PM \\
        Vic      & 523  & 20 & 2490  & 175     & 23-Jun-25 & 8:36 PM & 10:18 PM & 8:55 PM \\
        City     & 447  & 17 & 2143  & 100     & 21-Jun-25 & 8:28 PM & 10:37 PM & 8:55 PM \\
        West & 715  & 28 & 3459  & 50  & 02-Jun-25 & 4:38 PM & 10:21 PM & 8:45 PM \\
        \midrule
        \textbf{Total} & 2227 & 86 & 10671 &        &           &          &          &         \\
        \bottomrule
    \end{tabular}
\end{table*}
\section{Methodology}
\label{sec:methodology}

The key idea is to bridge the RGB–thermal modality gap by performing localization in a shared semantic domain. Both modalities are converted into semantically labeled structural representations and geometric consistency is enforced in a global 3D frame, enabling cross-modal alignment without appearance consistency. The overall pipeline consists of four stages:

\begin{enumerate}
\item[(A)] \textit{RGB semantic segmentation:} Daytime RGB imagery is segmented to produce dense semantic labels used to supervise cross-modal transfer and to construct a persistent semantic map.

\item[(B)] \textit{Thermal semantic segmentation:} A thermal segmentation model is trained using aligned RGB–thermal data, enabling nighttime thermal imagery to be converted into semantically labeled observations.

\item[(C)] \textit{Semantic map construction through reprojection and pruning:} A geo-referenced 3D map is reconstructed from daytime RGB data, semantic labels are reprojected into the 3D point cloud, and structural class boundaries are retained through edge-aware pruning to produce a semantically labeled 3D edge map.

\item[(D)] \textit{Nighttime semantic map-based localization:} During nighttime operation, segmented thermal observations are aligned with the semantic 3D map and the UAV translation is estimated by minimizing a symmetric semantic alignment objective.
\end{enumerate}

Stages (A)–(C) are performed offline to construct a semantically-labeled 3D map and to supplement training of the thermal semantic segmentation model, while Stage (D) performs map-relative localization using thermal imagery alone at inference time.

\subsection{RGB Semantic Segmentation}
\label{sect_rgb_seg}

A semantic segmentation network was trained on 78,162 aerial image patches from four datasets: FLAIR \cite{garioud_flair_nodate}, Semantic Drone \cite{Rahman_nodate}, UAVid \cite{lyu_uavid_2020}, and VDD \cite{cai_vdd_2025}. All images were standardized to $512\times512$, and classes were consolidated into localization-relevant categories using the FLAIR taxonomy as the primary reference due to its scale and annotation consistency. Dataset contributions were: FLAIR (54,188 patches), VDD (280 originals $\times$35 tiling), UAVid (70 originals $\times$30 tiling), and Semantic Drone (400 originals $\times$2 tiling). Augmentation was applied inversely to dataset size to balance representation across datasets.

Source labels were mapped into eight unified semantic classes: Animal, Building, Impervious Surface, Pervious Surface, Tree Vegetation, Low Vegetation, Water, and Vehicle. These classes correspond to persistent structural elements commonly observed in aerial environments and were selected to preserve semantic consistency while maximizing cross-dataset compatibility.

The model uses DeepLabV3+ \cite{ferrari_encoder-decoder_2018} with a ResNet \cite{he_deep_2016} backbone and MiT-B5 \cite{xie_segformer_2021} encoder pretrained on ImageNet \cite{russakovsky_imagenet_2015}. Training is performed using Adam and class-weighted cross-entropy, retaining the best validation checkpoint.

\subsection{Thermal Semantic Segmentation and RGB--Thermal Alignment}

This cross-modal teacher--student distillation is conceptually similar to HeatNet~\cite{vertens_heatnet_2020}, which transfers semantic supervision across RGB and thermal domains so that a model trained with daytime-aligned labels can generalize to nighttime thermal imagery. In our case, we use only daytime synchronized RGB--thermal pairs to generate aligned pseudo-labels (via the RGB teacher) for training, and the resulting thermal student is then deployed on nighttime thermal frames at test time.

\paragraph{Data collection and cross-modal supervision}
We collected 2,270 daytime RGB--thermal frame pairs using co-mounted RGB and thermal cameras across flights at 30--175\,m AGL. We first train an RGB semantic segmentation network (teacher) on RGB imagery from the aggregated dataset (Sec.~\ref{sect_rgb_seg}). We then use its predictions on the collected RGB--thermal pairs to supervise a thermal segmentation network (student) with the same architecture and identical class set (Fig.~\ref{fig:pipeline}), avoiding manual pixel-level annotation in thermal imagery.

\paragraph{RGB--thermal alignment}
To transfer labels across modalities, we register RGB pixels into the thermal image plane. Since the RGB and thermal sensors are rigidly mounted on the same gimbal with a small baseline relative to flight altitude, the mapping is approximately constant across the dataset; we therefore estimate a single global projective transform. Specifically, 120 corresponding feature points were manually annotated across representative paired frames and used to fit a homography $\mathbf{H}_{T\leftarrow R}\in\mathbb{R}^{3\times 3}$. For each paired sample, the RGB image is warped using $\mathbf{H}_{T\leftarrow R}$ (nearest-neighbor interpolation) and cropped to $512\times512$ to align the images and then is pushed through the teacher model to produce aligned pseudo-labels. These aligned masks provide supervisory targets for training the thermal student network.

Using the aligned pairs, thermal images are provided to the student as single-channel (grayscale) inputs. Pixels outside the overlapping field of view and cropping region are ignored during training.

\subsection{Semantic Map Construction via Reprojection and Pruning} 

Structure-from-Motion reconstruction from daytime RGB
flight, performed using RealityCapture \cite{reality_reality_2025} produces a geo-referenced colored point cloud $\mathcal{X}=\{(\mathbf{x}_w^i,\mathbf{c}^i)\}$,
where $\mathbf{x}_w^i \in \mathbb{R}^3$ is a 3D point in the world frame and $\mathbf{c}^i\in\mathbb{R}^3$ is its RGB color.

To assign semantic labels, each 3D point is projected into segmented RGB images in which it is visible. Given camera intrinsics $\mathbf{K}$ and extrinsics $(\mathbf{R}_{wc},\mathbf{t}_{wc})$, the projection operator $p(\cdot)$ maps $\mathbf{x}_w^i$ to pixel coordinates $(u,v)$. Occlusions are handled via z-buffering, retaining only the closest point per pixel. The per-view semantic label is read as $\mathrm{seg}_I(p(\mathbf{x}_w^i))$.

For points observed in multiple views, labels are fused via majority voting:
\begin{equation}
\hat{y}^i = \arg\max_{y\in\mathcal{Y}} 
\sum_{I\in \mathcal{V}_i}
\mathbf{1}\!\big[\mathrm{seg}_I(p(\mathbf{x}_w^i)) = y\big],
\end{equation}
where $\mathcal{V}_i$ denotes the set of images in which point $i$ is visible and $\mathcal{Y}$ is the semantic class set. The resulting semantically labeled map is $\mathcal{M}=\{(\mathbf{x}_w^i,\hat{y}^i)\}$.

To emphasize structural information relevant for alignment, the labeled cloud is voxelized and voxels exhibiting class transitions between neighboring cells are retained while large single-class regions are pruned. This produces a semantically labeled 3D edge map used for downstream registration (Figure~\ref{fig:pipeline}).

\begin{table*}[t]
    \centering
    \caption{Performance Comparison of RGB and Thermal Segmentation Models}
    \label{tab:rgb_vs_thermal}
    \begin{tabular}{llcccccccccc}
        \toprule
        \multicolumn{2}{c}{\textbf{Classes}} &
        \multicolumn{4}{c}{\textbf{RGB Model}} &
        \multicolumn{4}{c}{\textbf{Thermal Model}} &
        \multicolumn{2}{c}{\textbf{Pixels for Each Class (\%)}} \\
        \cmidrule(lr){1-2}\cmidrule(lr){3-6}\cmidrule(lr){7-10}\cmidrule(lr){11-12}
        \textbf{Name} & \textbf{Colour} &
        \textbf{Prec.} & \textbf{Rec.} & \textbf{F1} & \textbf{IoU} &
        \textbf{Prec.} & \textbf{Rec.} & \textbf{F1} & \textbf{IoU} &
        \textbf{Thermal Set} & \textbf{RGB Set} \\
        \midrule
        Building            & Yellow      & 91.8 & 93.0 & 92.4 & 85.9 & 67.9 & 79.7 & 73.3 & 57.9 & 8.9 & 2.7 \\
        Impervious Surface  & Grey        & 84.7 & 84.1 & 84.4 & 73.0 & 82.5 & 79.7 & 81.1 & 68.2 & 28.6 & 10.0 \\
        Pervious Surface    & Brown       & 92.3 & 91.9 & 92.1 & 85.4 & 69.2 & 56.7 & 62.3 & 45.3 & 4.3 & 19.3 \\
        Tree Vegetation     & Dark Green  & 90.5 & 91.2 & 90.9 & 83.2 & 86.6 & 88.3 & 87.4 & 77.6 & 26.7 & 23.1 \\
        Low Vegetation      & Light Green & 91.5 & 91.3 & 91.4 & 84.2 & 81.3 & 78.8 & 80.0 & 66.7 & 29.0 & 37.9 \\
        Water               & Blue        & 91.1 & 91.6 & 91.3 & 84.1 & 78.3 & 73.7 & 76.0 & 61.2 & 1.2 & 6.9 \\
        Vehicle             & Red         & 93.0 & 91.5 & 92.2 & 85.6 & 37.4 & 65.7 & 47.6 & 31.3 & 0.1 & 0.1 \\
        \midrule
            & \textbf{Average}              & 90.7 & 90.7 & 90.7 & 83.1 & 71.9 & 74.7 & 72.5 & 58.3 &      &      \\
        \bottomrule
    \end{tabular}
\end{table*}

\subsection{Nighttime Semantic Map-Based Localization}

Nighttime localization registers a segmented thermal image to the semantic 3D map $\mathcal{M}$. We assume camera intrinsics $\mathbf{K}$, camera-to-body extrinsics $(\mathbf{R}_{cb}, \mathbf{t}_{cb})$, and the body attitude estimate $\hat{\mathbf{R}}_{wb} \in SO(3)$ are available from onboard state estimation (IMU and Barometer). Similarly, the camera height $\hat{h}$ is assumed known. The only unknown is the planar translation
\begin{equation}
    \mathbf{t} = (t_x, t_y) \in \mathbb{R}^2.
\end{equation}


\paragraph{Semantic map projection} Let $(\mathbf{x}_w^i, y_i) \in \mathcal{M}$ denote a labeled 3D map point.
For a candidate translation $\mathbf{t}$, the point is transformed into the camera frame and projected into the image using the known camera model, yielding pixel coordinates $p_i(\mathbf{t}) \in \mathbb{R}^2$. The set of projected semantic points is
\begin{equation}
    \Pi(\mathbf{t}) = \{(p_i(\mathbf{t}),y_i)\}.
\end{equation}

\paragraph{Thermal segmentation edges}
Let the segmented thermal image yield class-specific edge sets
\begin{equation}
    \mathcal{E}_k = \{e_j \in \mathbb{R}^2 \mid \text{edge pixel predicted as class } k\},    
\end{equation}

with $\mathcal{E}=\bigcup_{k=1}^K \mathcal{E}_k$.
Each edge pixel $e_j$ has predicted label $y_j$. Localization is formulated as minimizing geometric inconsistency between projected map structure $\Pi(\mathbf{t})$ and segmentation edges $\mathcal{E}$.
\paragraph{Bidirectional Semantic Alignment Loss} For any point $\mathbf{q} \in \mathbb{R}^2$ and finite set $S \subset \mathbb{R}^2$, define the point-to-set distance
\begin{equation}
    d(\mathbf{q},S) = \min_{\mathbf{s} \in S} \| \mathbf{q} - \mathbf{s} \|_2.
\end{equation}

Localization is formulated as minimizing a symmetric semantic Chamfer objective:
\begin{equation}
    \mathcal{L}(\mathbf{t}) = \lambda_f \mathcal{L}_f(\mathbf{t}) + \lambda_r \mathcal{L}_r(\mathbf{t}),
\end{equation}
with weights $\lambda_f, \lambda_r > 0$. The forward term enforces that projected semantic map boundaries coincide with observed segmentation edges of the same class:
\begin{equation}
\mathcal{L}_f(\mathbf{t}) =
\frac{1}{|\Pi(\mathbf{t})|}
\sum_{(p_i(\mathbf{t}), y_i) \in \Pi(\mathbf{t})}
\rho_\delta\!\left(
d\!\left(p_i(\mathbf{t}), \mathcal{E}_{y_i}\right)
\right),
\end{equation}
where $\rho_\delta(\cdot)$ denotes the Huber penalty. The reverse term enforces that observed edges are explained by projected map geometry:
\begin{equation}
\mathcal{L}_r(\mathbf{t}) =
\frac{1}{|\mathcal{E}|}
\sum_{(e_j, \hat y_j) \in \mathcal{E}}
\rho_\delta\!\left(
d\!\left(\mathbf{e}_j, P_{\hat y_j}(\mathbf{t})\right)
\right),
\end{equation}
where $P_k(\mathbf{t}) = \{p_i(\mathbf{t}) \mid y_i = k\}$. Together, $\mathcal{L}_f$ and $\mathcal{L}_r$ approximate a symmetric semantic Chamfer distance, enforcing mutual geometric consistency between projected map structure and observed thermal edges.

\paragraph{Confusion-Aware Semantic Matching}

Thermal segmentation exhibits systematic inter-class ambiguity. Let
\begin{equation}
C_{y,k} = \Pr(\hat y = k \mid y),
\qquad
\sum_{k=1}^K C_{y,k} = 1,
\end{equation}
denote the empirical confusion matrix computed on the aligned daytime RGB--thermal supervision set.

We replace hard same-class matching with confusion-marginalized matching. The forward term becomes
\begin{equation}
\mathcal{L}_f(\mathbf{t}) =
\frac{1}{|\Pi(\mathbf{t})|}
\sum_{(p_i(\mathbf{t}), y_i) \in \Pi(\mathbf{t})}
\sum_{k=1}^K
C_{y_i,k}\;
\rho_\delta\!\left(
d\!\left(p_i(\mathbf{t}), \mathcal{E}_k\right)
\right),
\end{equation}
and the reverse term is weighted analogously. This marginalization reduces brittleness under nighttime degradation by allowing alignment to fall back to statistically likely substitute classes.

\paragraph{Optimization} The translation estimate is obtained as
\begin{equation}
\mathbf{t}^\star = \arg\min_{\mathbf{t} \in \mathcal{T}} \mathcal{L}(\mathbf{t}),
\end{equation}
where $\mathcal{T} = [-r,r] \times [-r,r]$ is a bounded search region (in metres) centered at the onboard navigation prior. In our experiments, we select $r = 30$\,m. Because $\mathcal{L}(\mathbf{t})$ is nonconvex and involves point-to-set minima, we employ a coarse-to-fine discrete search. The objective is evaluated on a sequence of grids with progressively decreasing spacings $\Delta_0 > \Delta_1 > \dots > \Delta_S$, each stage re-centered around the best candidate from the previous level. This provides deterministic computation cost and robust convergence under segmentation noise and partial map visibility.

\begin{figure*}[t]
    \centering
    \includegraphics[width=1\textwidth]{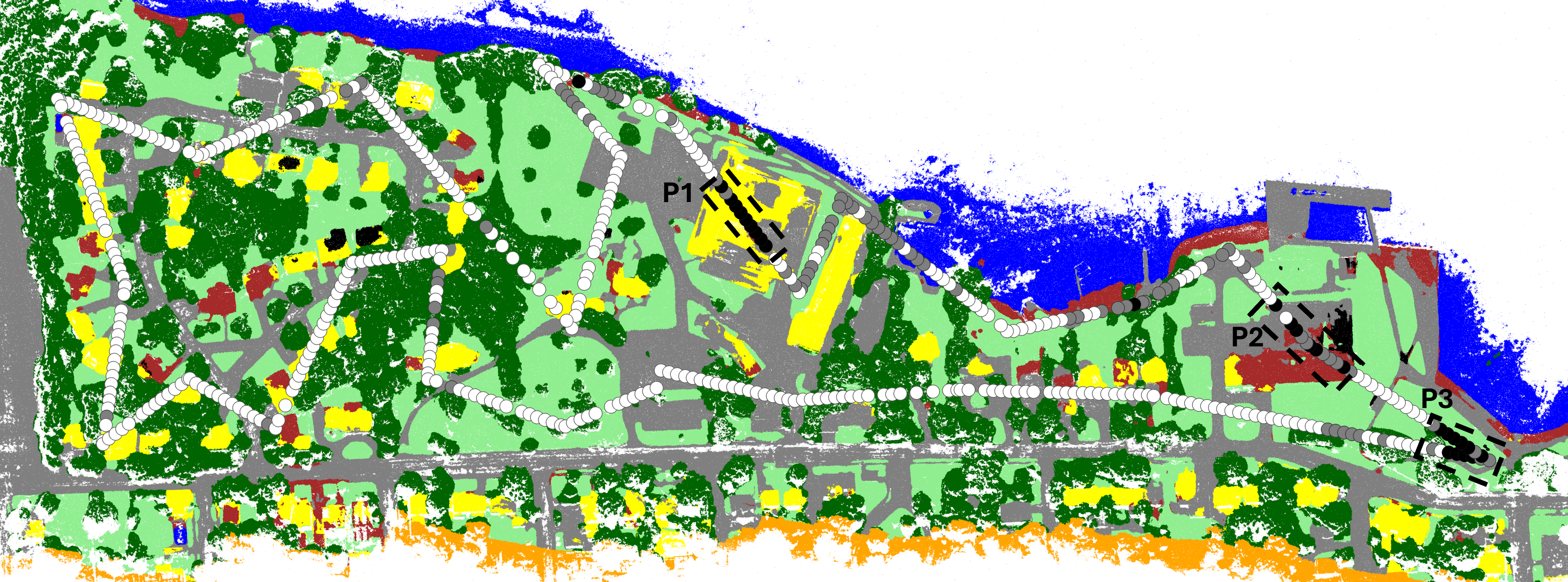}
    \caption{\small Pier flight trajectory overlaid on the semantically labeled global map. Localization error relative to RTK ground truth is color-coded along the path: white indicates high-accuracy estimates (0–2\,m), grey denotes moderate error (2–5\,m), and black highlights failure regions exceeding 5\,m. Regions P1–P3 mark regions where large errors are concentrated. The figure illustrates that localization failures are confined to specific scene configurations.}
    \label{fig:pier_2d}
\end{figure*}
\section{Results and Evaluation}
\subsection{Hardware Configuration}
All flight trials and data collection were conducted using a DJI Matrice 350 RTK UAV equipped with a Zenmuse H20T multi-sensor payload. The wide-angle RGB and thermal (40.6\textdegree{} DFOV, LWIR 8--14\,$\mu$m) sensors were used for data acquisition. A summary of flight sites, trajectory lengths, altitudes, and dataset statistics is provided in Table~\ref{tab:flight_paths}.

Flights were preplanned at fixed altitudes that varied by site to evaluate robustness across operating conditions. RTK positioning was used during both daytime and nighttime flights to geo-reference the map and to generate ground-truth trajectories for quantitative evaluation. Importantly, RTK measurements were used solely for map construction and performance benchmarking and were not provided as input to the nighttime localization pipeline.

\subsection{Semantic Segmentation Performance}

Semantic segmentation results for the RGB teacher and thermal student models are summarized in Table~\ref{tab:rgb_vs_thermal}. The RGB model achieved an average IoU of 83.1\%, with all classes exceeding 73\% IoU. 

The thermal model obtained an average IoU of 58.3\%, with higher performance on structurally dominant classes such as tree vegetation (77.6\%), low vegetation (66.7\%), and impervious surfaces (68.2\%). Buildings achieved 57.9\% IoU, while vehicles remained the most challenging class (31.3\%) due to low representation and weak thermal contrast.

Despite lower overall accuracy, the thermal model preserved the primary structural classes used for localization; impervious surfaces, vegetation, and buildings, which together account for over 90\% of thermal pixels. These results indicate that segmentation quality is sufficient to support semantic alignment and downstream localization.

The West environment, flown at the lowest altitude (50\,m AGL), exposed an observability limit of the method: images were dominated by single semantic classes, providing insufficient edge structure for reliable alignment. This confirms that semantic edge diversity is a necessary condition for the reprojection-based approach.

\subsection{Semantic Point Cloud Construction}

Semantic labeling of the reconstructed point cloud was evaluated qualitatively. Reprojection-based label fusion produced consistent semantic structure across views, improving robustness compared to single-image inference. 

Pruning reduced the point count to approximately 4\% of its original size while preserving structural edges critical for localization. The City dataset was reduced from 64M to 3M points, significantly lowering computational cost without degrading alignment quality (Figure ~\ref{fig:small_diagram}).

\begin{table*}[t]
\centering
\caption{Bias-corrected horizontal localization accuracy across datasets.}
\label{tab:localization_accuracy}
\begin{tabular}{lcccccccc}
\hline
Dataset & Frames & RMSE$_x$ & RMSE$_y$ & RMSE$_{2D}$ & Median$_{2D}$ & P75$_{2D}$ & \%$<2$m & \%$>5$m \\
\hline
City Park & 402 & 1.151 & 1.526 & 2.126 & 1.499 & 2.162 & 70\% & 5\% \\
Vic Park  & 482 & 1.472 & 1.779 & 2.589 & 1.815 & 2.666 & 56\% & 8\% \\
Pier      & 489 & 1.265 & 1.020 & 1.831 & 1.334 & 1.884 & 55\% & 6\% \\
\hline
Total     & 1373 & 1.305 & 1.435 & 2.184 & 1.519 & 2.299 & 68\% & 6\% \\
Total (Gated 8000, 90\%)  & 1247 & 1.216 & 1.357 & 2.053 & 1.493 & 2.186 & 70\% & 5\% \\
Total (Gated 10000, 70\%) & 1102 & 1.171 & 1.309 & 1.979 & 1.465 & 2.141 & 72\% & 5\% \\
\hline
\end{tabular}
\end{table*}

\begin{table}[t]
\centering
\caption{Localization error versus edge-point count bin.}
\label{tab:edge_bins}
\begin{tabular}{lccc}
\hline
Edge Pixels (bin) & Std. Dev. (m) & Mean (m) & $N$ \\
\hline
1749--7248    & 3.847 & 3.799 & 88  \\
7249--12748   & 2.773 & 2.446 & 487 \\
12749--18248  & 2.171 & 1.953 & 586 \\
18249--23748  & 2.167 & 1.611 & 184 \\
23749--29248  & 0.434 & 1.131 & 28  \\
\hline
Total         & 2.566 & 2.184 & 1373 \\
\hline
\end{tabular}
\end{table}

\subsection{Localization Performance and Analysis}

Localization accuracy was evaluated against RTK ground truth (reported error 1--3\,cm) across all datasets. Horizontal localization error is computed in the $x$--$y$ plane and reported as $\mathrm{RMSE}_{2D}$. Over 1{,}373 frames spanning 6.5\,km of nighttime flight, the system achieved a combined $\mathrm{RMSE}_{2D}$ of 2.18\,m with a median horizontal error of 1.52\,m (Table~\ref{tab:localization_accuracy}). Overall, 68\% of frames localized within 2\,m, while 6\% exceeded 5\,m.

Performance varied by environment. Pier achieved the lowest $\mathrm{RMSE}_{2D}$ (1.83\,m), followed by City (2.13\,m). Vic showed higher error (2.59\,m), likely due to increased semantic ambiguity. Because alignment is performed over a bounded spatial region and relies on edge-based reprojection, moderate attitude errors primarily manifest as consistent spatial shifts in projected map structure. In practice, attitude estimates from IMU/VIO pipelines are significantly more accurate than meter-level translation errors over short time horizons, making translation the dominant uncertainty term in the global correction step. Typical drift in monocular VIO on kilometer-scale trajectories can exceed 1–3\% of the distance traveled, corresponding to 20–60 m over 2 km without loop closure. The proposed method reduces global error to a meters-level accuracy using thermal imagery alone.

An ablation study assessed the influence of key framework components. \paragraph{Bidirectional alignment}  
The symmetric formulation combining forward (projection-to-edge) and reverse (edge-to-projection) terms proved critical for stable convergence. Using either term alone increased both mean error and variance. The forward-only formulation showed susceptibility to sparse projected-point attraction despite sharper local alignment in well-constrained regions. The combined formulation reduced error dispersion by approximately 1.5\,m across datasets and consistently produced the most stable convergence behavior. \paragraph{Huber loss parameters} 
Huber thresholds were evaluated across a range of values. A setting of $\delta = 2$\,px with a 5\,px clamp provided the best balance between sensitivity to small misalignments and robustness to outliers. All reported results use this configuration. \paragraph{Confusion-aware weighting}  
Incorporating class-confusion statistics into the symmetric loss reduced mean horizontal error by 12.1\% and standard deviation by 17.5\% across datasets, while median error improved by 2\%. Improvements were strongest in semantically ambiguous environments: City mean error decreased from 2.52\,m to 2.13\,m (15.5\%), and Vic from 3.14\,m to 2.59\,m (17.5\%). These results indicate that confusion-aware weighting primarily suppresses large-error outliers.
\textit{Runtime performance.}  
Runtime on an NVIDIA GeForce RTX 4070 GPU averages 0.95\,s per update, sufficient for periodic global correction when coupled with high-rate VIO.

\paragraph{Effect of Semantic Edges}

Localization accuracy was strongly correlated with available semantic edge evidence. Applying an evidence threshold of 8{,}000 edge points retained approximately 90\% of frames ($N=1{,}247$) while improving $\mathrm{RMSE}_{2D}$ from 2.18\,m to 2.05\,m by effectively reducing outliers (Table~\ref{tab:edge_bins}). These results indicate that semantic reprojection localization exhibits a structural observability dependence: when semantic edge density is low, the alignment objective becomes weakly conditioned and multiple translations yield similar loss values.

\begin{figure}[t]
    \centering

    \begin{subfigure}{0.98\linewidth}
        \centering
        \includegraphics[width=\linewidth]{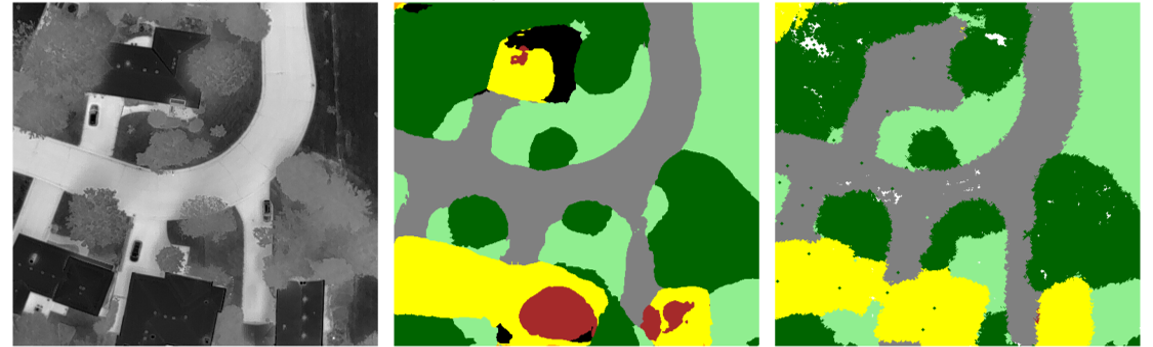}
        \caption{Top 10\% performance. Strong semantic detail and clear structural boundaries, resulting in 0.4\,m $\mathrm{RMSE}_{2D}$.
 }
        \label{fig:qual_top10}
    \end{subfigure}

    \vspace{4pt}

    \begin{subfigure}{0.98\linewidth}
        \centering
        \includegraphics[width=\linewidth]{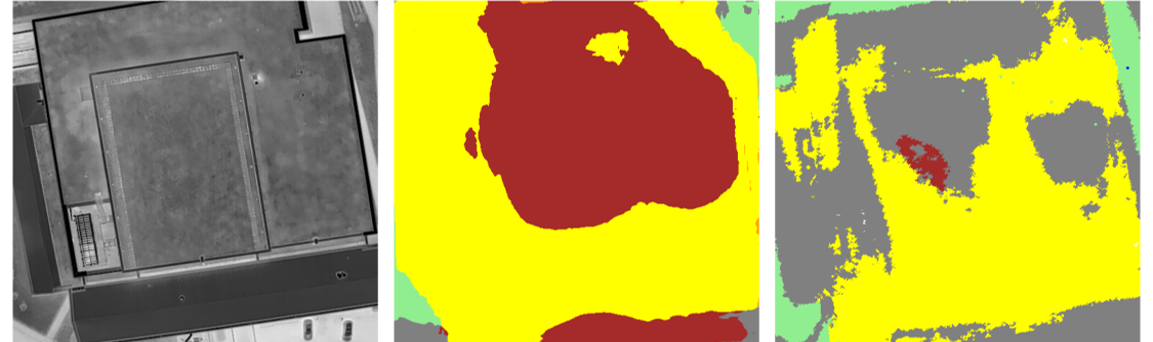}
        \caption{Region P1. Large building dominates the view, limiting semantic diversity and yielding 8.8\,m $\mathrm{RMSE}_{2D}$.
}
        \label{fig:qual_p1}
    \end{subfigure}

    \vspace{4pt}

    \begin{subfigure}{0.98\linewidth}
        \centering
        \includegraphics[width=\linewidth]{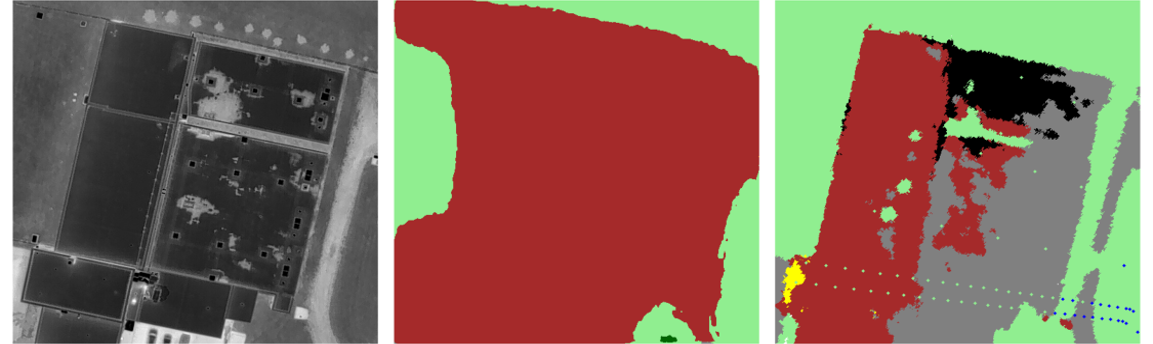}
        \caption{Region P2. Large building dominates the view, leading to incorrect classification and yielding 11.0\,m $\mathrm{RMSE}_{2D}$}
        \label{fig:qual_p2}
    \end{subfigure}

    \vspace{4pt}

    \begin{subfigure}{0.98\linewidth}
        \centering
        \includegraphics[width=\linewidth]{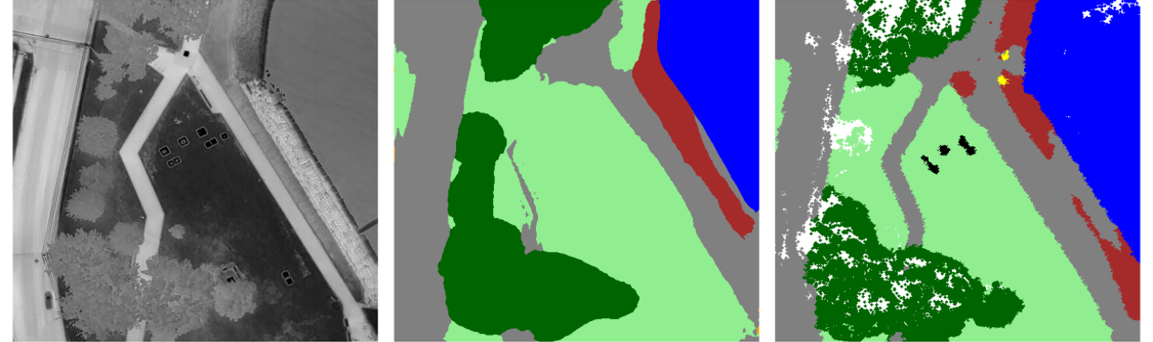}
        \caption{Region P3. Ambiguity in the water class induces both vertical and horizontal alignment shifts despite otherwise consistent semantic classification, resulting in 5.5\,m $\mathrm{RMSE}_{2D}$.}
        \label{fig:qual_p3}
    \end{subfigure}

    \caption{\small Qualitative localization examples shown as row-wise triplets (Left: thermal image. Middle: thermal semantic segmentation. Right: reprojected RGB semantic map) from Pier failure regions. Reprojections are computed using RTK ground-truth poses to isolate segmentation effects from localization error.}
    \label{fig:qual_localization_strips}
\end{figure}

\paragraph{Spatial Distribution of Large-Error Estimates} We categorize large-error events into three classes: (i) low semantic edge density, (ii) systematic class confusion (e.g., water/impervious), (iii) map–environment mismatch (e.g., vegetation changes). Fig.~\ref{fig:pier_2d} highlights all frames exceeding 5\,m horizontal error in black along a representative Pier nighttime trajectory and Fig.~\ref{fig:qual_localization_strips} shows corresponding semantic segmentation examples for three challenging regions P1--P3 marked on Fig.~\ref{fig:pier_2d}. Across the Pier trajectory ($N=489$), only 27 frames (5.5\%) exceed 5\,m error, and 25 occur within three compact regions (P1--P3) corresponding to areas of limited structural diversity, large homogeneous surfaces or ambiguous class boundaries. Specifically P3 is shifted due to inaccurate shorelines resulting from its under representation in the thermal training (Table \ref{tab:rgb_vs_thermal}).




\section{Conclusion and Future Work}

This work presents a map-relative nighttime UAV localization framework that bridges the RGB–thermal modality gap through semantic reprojection in a global 3D frame. By aligning segmented thermal observations with a semantically labeled 3D map constructed from daytime RGB data, the method replaces appearance-based matching with structural consistency constraints. Across 6.5\,km of real-world nighttime flight (1{,}373 frames), the system achieves an $\mathrm{RMSE}_{2D}$ of 2.18\,m relative to RTK ground truth. Performance analysis shows that symmetric bidirectional alignment and confusion-aware weighting significantly reduce error dispersion and suppress large-error outliers. Furthermore, localization performance is strongly correlated with semantic edge density, consistent with the structural observability analysis in Section~\ref{sec:methodology}.

The current work focuses on the global correction module; onboard integration with high-rate thermal odometry remains as future work. Future improvements are expected to arise primarily from advances in segmentation fidelity. Frames with limited structural diversity or ambiguous class boundaries dominate failure cases, suggesting that improved boundary accuracy, altitude-balanced training data, and better representation of under-sampled classes (e.g., water) could directly enhance localization robustness. The current implementation is also constrained by the narrow thermal field of view and the cropped $512\times512$ input resolution, which limit the observable context for alignment.

\bibliographystyle{IEEEtran}
\bibliography{ICUAS3}

\end{document}